\pdfoutput=1
\documentclass{article}
\usepackage{ijcai18}

\usepackage{times}
\usepackage{xcolor}
\usepackage{soul}
\usepackage[small]{caption}
\usepackage{CJK}
\usepackage{amsmath}
\usepackage{tabularx} 
\usepackage{booktabs}  
\usepackage{threeparttable}  
\usepackage{graphicx}

\title{Mask-aware Photorealistic Face Attribute Manipulation}

\author{
Ruoqi Sun$^1$, 
Chen Huang$^2$, 
Jianping Shi$^3$, 
Lizhuang Ma$^1$
\\ 
$^1$ Shanghai Jiao Tong University \\
$^2$ Carnegie Mellon University\\
$^3$ Beijing Sensetime Tech. Dev. Co., Ltd \\
ruoqisun7@sjtu.edu.cn,
chenh2@andrew.cmu.edu,
shijianping@sensetime.com,
ma-lz@cs.sjtu.edu.cn,
}
\begin{document}

\maketitle

\begin{abstract}

The task of face attribute manipulation has found increasing applications, but still remains challenging with the requirement of editing the attributes of a face image while preserving its unique details. In this paper, we choose to combine the Variational AutoEncoder (VAE) and Generative Adversarial Network (GAN) for photorealistic image generation. We propose an effective method to modify a modest amount of pixels in the feature maps of an encoder, changing the attribute strength continuously without hindering global information. Our training objectives of VAE and GAN are reinforced by the supervision of face recognition loss and cycle consistency loss for faithful preservation of face details. Moreover, we generate facial masks to enforce background consistency, which allows our training to focus on manipulating the foreground face rather than background. Experimental results demonstrate our method, called Mask-Adversarial AutoEncoder (M-AAE), can generate high-quality images with changing attributes and outperforms prior methods in detail preservation.

\end{abstract}

\section{Introduction}

The task of face attribute manipulation is to edit the face attributes shown in an image, e.g., hair color, facial expression, age and so on. It has a wide range of applications, such as data augmentation and age-invariant face verification~\cite{park2010age,Chi2017Temporal}. Essentially, this is an image generation problem. But unlike the style translation task~\cite{Gatys2016Image,Li2017Demystifying}, the attribute manipulation one is more challenging due to the requirement of only modifying some image features while keeping others unchanged (including the image background).

With the advent of generative adversarial networks (GANs)~\cite{goodfellow2014generative}, the quality of generated images improves over time. The family of GAN methods can be mainly divided into two categories: one with noise input~\cite{Mirza2014Conditional,yang2017lr} and the other conditioned on an input images~\cite{johnson2016perceptual,Wang2017Tag,Choi2017StarGAN}. Our method falls into the second category, with the aim to change the face attributes in the input image with highly-preserved details. 

One simple option to achieve this goal is to use the conditional GAN framework~\cite{Mirza2014Conditional,Zhang2017Age}, which concatenates the input image with a one-hot attribute vector to encode the desired manipulation. However, such global transformation cannot either guarantee facial detail preservation, or make a continuous change in the attribute strength. Another option is to directly learn the image-to-image translation along attributes. CycleGAN~\cite{zhu2017unpaired} learns such translation rule from unpaired images with a cycle consistency constraint. The recent UNIT method~\cite{Liu2017Unsupervised} uses generative adversarial networks (GANs) and variational autoencoders (VAEs) for robust modelling of different image domains. Then the cycle consistency constraint is also applied to learn domain translation effectively.~\cite{Shen2017Learning} proposed to only learn the residual image before and after attribute manipulation by using two transformation networks, one for attribute manipulation and the other for its dual operation.

The above methods share one common drawback --- there exists no mechanisms to keep the unique facial traits while editing attributes. Most likely we will observe changed attributes with lost personal details.~\cite{Zhang2017Age} provided a partial remedy by feeding the face images before and after attribute manipulation into a face recognition network and penalizing their feature distance. This is essentially one way to preserve facial identify information. However, it may still change the non-targeted features beyond identity or other parts of the image (e.g., background), which is not visually pleasing. We specially note the importance of keeping the background unchanged since it is often observed to be changed along with the foreground face. This suggests some face attribute manipulation efforts are wasted in the irrelevant regions. Pasting the original background around the manipulated face with a face mask would not be the solution, because the two parts can be drastically incompatible.
%can only process the low-resolution images or face images without background.

In this paper, we learn to simultaneously manipulate the target attributes of a face image and keep its background untouched. Our method is based on the VAE-GAN framework~\cite{Zhang2017Age,Liu2017Unsupervised} for strong modeling of photorealistic images. We propose an effective method to modify a minimum number of feature map pixels from our encoder. This allows us to maximally preserve the global image information and also change the strength of target attributes continuously. To avoid loss of the unique facial details during attribute editing, we attach to the VAE-GAN objectives additional face recognition loss and cycle consistency loss (to ensure image consistency after two inverse manipulations). Furthermore, we mask out image backgrounds to coherently penalize their difference before and after face attribute manipulation. We call our method as Mask-Adversarial AutoEncoder (M-AAE) and support its efficacy by extensive experiments. 

In summary, the contributions of this paper are as follows:
\begin{itemize}
\item We present an effective method to modify a modest amount of pixels in our learned feature maps to realize continuous manipulation of face attribute.
\item We propose a Mask-Adversarial AutoEncoder (M-AAE) training objective to ensure faithful facial detail preservation as well as background consistency.
\item The proposed method demonstrates state-of-the-art performance in photorealistic attribute manipulation.
\end{itemize}

\section{Related Work}

\paragraph{Face attribute manipulation}

Most methods of face attribute manipulation are based on generative models. There are two main groups of these methods: the group with extra input vector, and the group that directly learn the image-to-image translation along attributes. The first group often takes an attribute vector as the guidance for manipulating the desired attribute. The CAAE method~\cite{Zhang2017Age} concatenates the one-hot age label with latent image features to be fed into the generator for age progression purposes. StarGAN~\cite{Choi2017StarGAN} takes the one-hot vector to represent domain information for "domain transfer". However, such global transformation based on external code usually cannot well preserve the facial details after attribute manipulation. The second group of methods only operate in image domains and learn the image-to-image translation directly. The CycleGAN~\cite{zhu2017unpaired} and UNIT method~\cite{Liu2017Unsupervised} are such examples, supervised by a cycle consistency loss that requires the manipulated image can be mapped back to the original image.~\cite{Shen2017Learning} further proposed to only learn the residual image before and after attribute manipulation, which can be easier and lead to higher-quality image prediction. Unfortunately, these methods still have difficulty of manipulating the target attribute while keeping others unchanged.

\noindent
{\bf VAE and GAN} The Variational AutoEncoder (VAE)~\cite{Kingma2014Auto} and Generative adversarial network (GAN)~\cite{goodfellow2014generative} are the backbone for image generation tasks nowadays, such as image synthesis~\cite{Radford2015Unsupervised,zhao2017tuch,yang2017lr} and image translation~\cite{Kim2017Learning,zhu2017unpaired,Liu2017Unsupervised}. In VAE, the encoder maps images into a latent feature space which is then mapped back to the image domain through a decoder. The latent space contains the global features extracted for input images. The more recent GAN consists of the generator and the discriminator networks to play a min-max game. Specifically, the generator tries to produce synthesized images to fool the discriminator that distinguishes the synthesized images from real ones. GAN-based methods have shown remarkable results in image generation, and many improvements followed up. DCGAN~\cite{Radford2015Unsupervised} trains stable in a purely convolutional setting, while CGAN~\cite{Mirza2014Conditional} generates visually compelling images conditioned on extra input like class labels. CycleGAN~\cite{zhu2017unpaired} and UNIT method~\cite{Liu2017Unsupervised} introduce a cycle consistency loss to learn between any image domains with even unpaired images. There is a recent trend to combine the GAN with a VAE for robust image modeling. For example,~\cite{Larsen2016Autoencoding} combined GAN and VAE by collapsing the VAE decoder and GAN generator into one. One can tweak the generated images by manipulating features in the latent feature space. Such joint VAE-GAN model is also applied in the works of~\cite{Zhang2017Age,Liu2017Unsupervised} for image translation. This paper uses the VAE-GAN model for face attribute manipulation, and proposes a working method to modify latent VAE features so as to change facial attributes but not irrelevant details.

 \begin{figure*}
	\begin{center}
			\includegraphics[width=1.7\columnwidth]{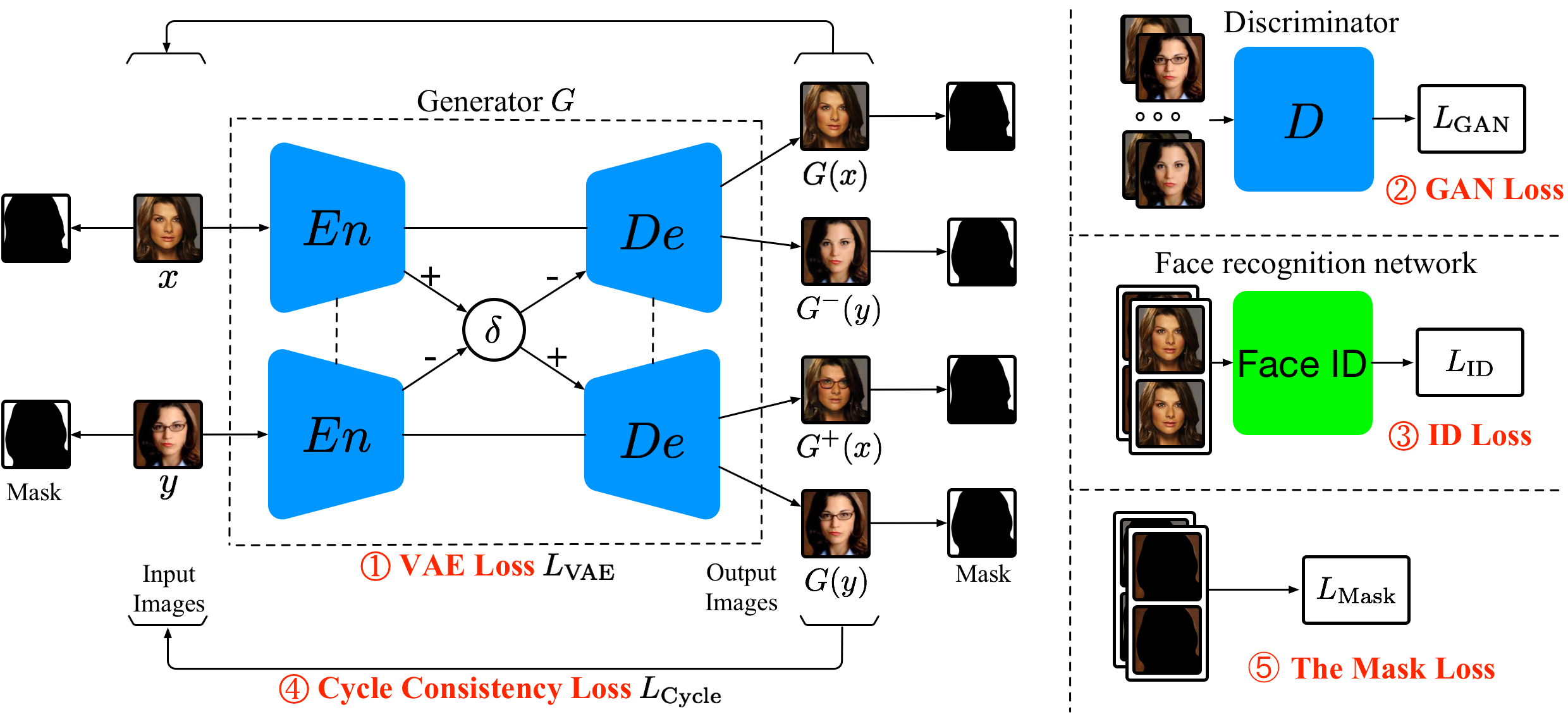}
	\end{center}
	\caption{Framework of the proposed Mask-Adversarial AutoEncoder (M-AAE) method. The encoder-decoder $De(En(x))$ of VAE for input image $x$ is treated as generator $G(x)$ of GAN, whose discriminator $D(\cdot)$ tells fake from real. We manipulate attributes by modifying the encoded features $En(x)$ by a relative value $\pm \delta$, and train using image pairs with opposite face attributes. Our training is supervised by 5 loss functions to both preserve facial details and ensure background consistency (see text for details). We test only using the generator $G(\cdot)$.}
	\label{Fig.1}
\end{figure*}

\section{Methodology}

Our goal is to manipulate the attribute of an input face image and generate a new one,~e.g.,~to change the hair color from black to yellow. The difficulty lies in the generation of photorealistic as well as faithful face images,~i.e.,~the generated image should look real and have its unique details preserved including the background. We propose a Mask-Adversarial AutoEncoder (M-AAE) method to address these challenges, as will be detailed as follows.

\subsection{Framework Overview}

Our M-AAE method is based on the VAE-GAN framework, as shown in Fig.~\ref{Fig.1}. The encoder-decoder $De(En(x))$ of VAE for input image $x$ is treated as GAN's generator $G(x)$. The discriminator $D(\cdot)$ of GAN tells the generated image $G(x)$ apart from real images. To manipulate attributes of input image $x$, we design a simple but effective mechanism to uniformly modify the encoded features $En(x)$ by a relative value $\pm \delta$, which is fed into the decoder to control the attribute strength present in output $G^+(x) / G^-(x)$.

\noindent
{\bf Training process} Besides training with the VAE and GAN loss functions, we also use the face recognition loss and cycle consistency loss for faithful preservation of face details. The face recognition module extracts features from images before and after attribute manipulation, and penalizes their feature discrepancy to preserve identity information. While the cycle consistency loss aims to preserve other unique facial information by penalizing the difference between input image $x$ and the generated image after two inverse attribute transformations $G^+(x)$ and $G^-(x)$. To ensure background consistency, we further generate facial masks to penalize the background difference between input $x$ and output $G(x)$.

\noindent
{\bf Testing process} We simply feed the input image $x$ through our generator $G(x)=De(En(x))$, changing the relative attribute strength $\delta$ in the latent features $En(x)$.

\subsection{Attribute Manipulation in Encoded Features}

To manipulate face attributes, rather than take a one hot attribute vector as in~\cite{Zhang2017Age,Choi2017StarGAN}, we choose to modify the latent features in our encoder to be able to continuously change the attribute strength. One intuitive way is to uniformly increase or decrease the responses of the entire feature map by a relative value $\delta$. We empirically observed a global change of image tone by doing this. Instead, we propose to only modify a minimum number of feature map pixels whose receptive field covers the whole image in image domain. Fig.~\ref{Fig.2} illustrates how to find such minimum pixels at the top feature layer recursively from bottom layer. In this way, the image-level manipulation can be operated efficiently with modest feature modification. More importantly, we will avoid a huge loss of image information. Our experiments will show our efficacy in information preservation during attribute manipulation.

In practice, the relative value $\delta$ is chosen as half the value range of the feature map pixels for reversing one particular attribute ($\delta \approx 5$ in our scenario). Then such modified features are fed into the decoder to generate output image $G^+(x)$ or $G^-(x)$ with strengthened or weakened attribute.

\subsection{Learning of Mask-Adversarial AutoEncoder}

\paragraph{VAE loss}

The VAE consists of an encoder that maps an image $x$ to a latent feature $z\sim En(x)=q(z|x)$ and a decoder that maps $z$ back to image space $x'\sim De(z)=p(x|z)$. The VAE regularizes the encoder by imposing a prior over the latent distribution $p(z)$, where $z\sim \mathcal{N}(0,I)$ is often assumed to have a Gaussian distribution. VAE also penalizes the reconstruction error between $x$ and $x'$, and has loss function: 
    \begin{equation}
       \begin{split}
           \mathcal{L}_{\text{VAE}}&=\lambda_1\text{KL}(q(z|x)||p(z))\\&-\lambda_2{E_{x\sim p_{\text{data}}(x)}}[\log~p(x'|x)],
        \end{split}
        \label{eq1}
    \end{equation}
where $\lambda_1$ and $\lambda_2$ balance the prior regularization term and reconstruction error term, and $\text{KL}$ is the Kullback-Leibler divergence. The reconstruction error term is actually equivalent to the $L1$ norm between $x$ and $x'$, since we assume $p(x|z)$ has a Laplacian distribution.

\paragraph{GAN loss}

The GAN loss is introduced to improve the photorealistic quality of the generated image. Since the encoder-decoder of VAE is treated as the GAN generator, we use the input image $x$ and generated image $G(x)$ from VAE as the real and fake images for discriminative training. The GAN loss function is as follows:
    \begin{equation}
       \begin{split}
           \mathcal{L}_{\text{GAN}}&={E_{x\sim p_{\text{data}}(x)}}[\log~D(x)]\\&+{E_{x\sim p_{\text{data}}(x)}}[\log~(1-D(G(x)))]. 
        \end{split}
    \end{equation}
The weights of the generator and discriminator are updated alternatively in the training process. 

\begin{figure*}[!t]
	\begin{center}
       \includegraphics[width=2.0\columnwidth]{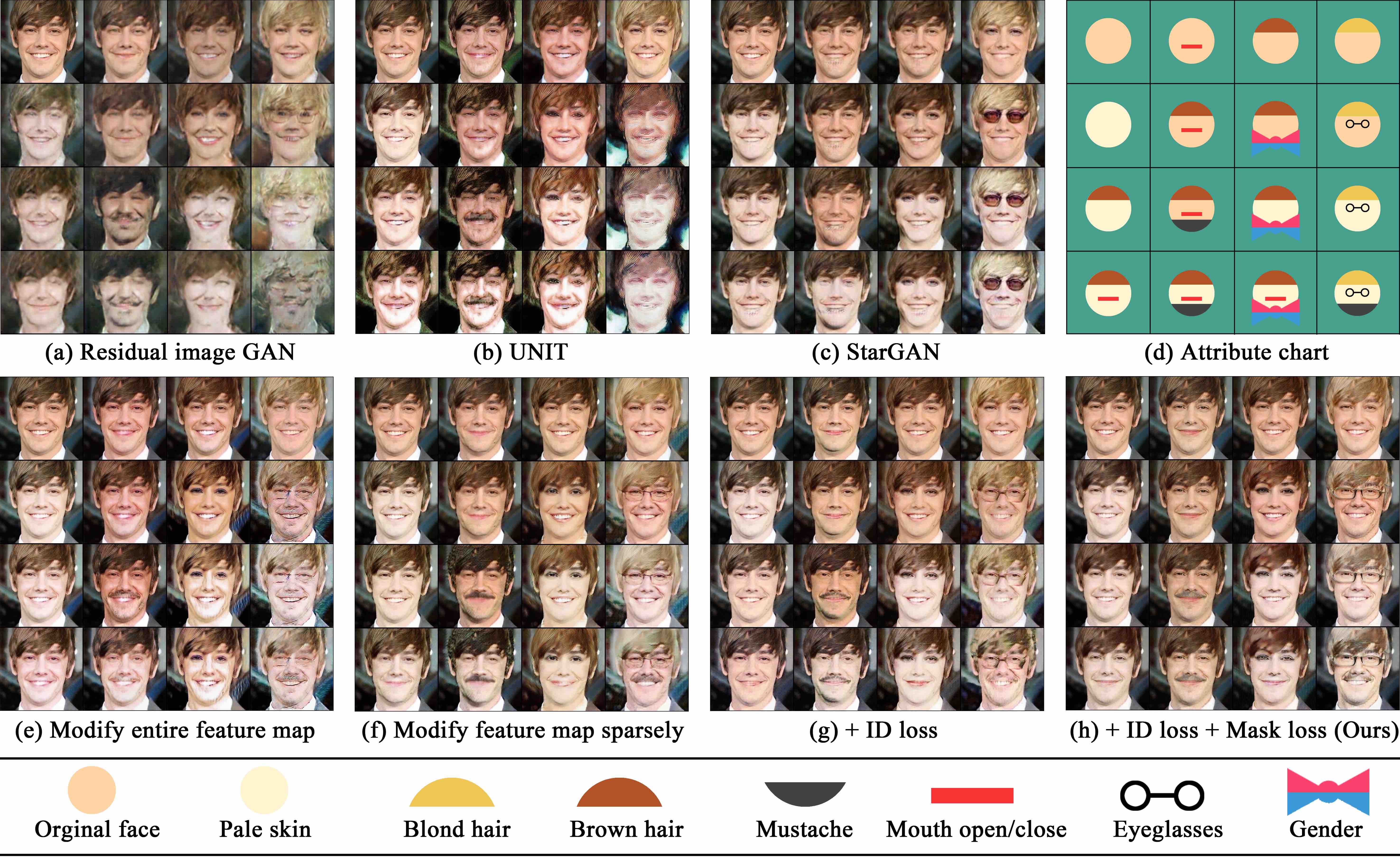}
	\end{center}
	\caption{Facial attribute manipulation results with 7 attributes on CelebA dataset. We compare with the state-of-the-art residual image GAN [Shen and Liu, 2017], UNIT [Liu \textit{et al.}, 2017] and StarGAN [Choi \textit{et al.}, 2017] (first row) and our various baselines (second row). For each method, the results are shown for the manipulation of corresponding attributes in the attribute chart.}
	\label{Fig.3}
\end{figure*}

\paragraph{ID loss}

To make the generated image photorealistic is not enough for face attribute manipulation. We can imagine an extreme case where one perfectly realistic generated image does not keep any unique traits about the face - it simply does not look alike the original face at all. This is not acceptable for faithful face manipulation. To preserve personal information as much as possible, we use a face recognition network~\cite{Parkhi2015Deep} to penalize the shift of face identity, which is one of the most important facial features to consider. Concretely, we extract identify features from images before and after attribute manipulation, and enforce them close to each other. The ID loss function is then defined as:
\begin{equation}
   	 \mathcal{L}_{ID} = \| F_{\text{ID}}(x) - F_{\text{ID}}(G(x)) \|^2,
\end{equation}
where  $F_{ID}(\cdot)$ is the feature extractor from the face recognition network.

\paragraph{Cycle consistency loss}

We still want to keep those facial charecteristics beyond identity after manipulation of the target attribute. Since it is hard to keep track of those charecteristics that have no supervision, we follow the idea of self-supervision in~\cite{zhu2017unpaired,Liu2017Unsupervised}. Specifically, we impose the cycle consistency constraint along the dimension of attribute. We apply two inverse transformations $G^+(\cdot)$ and $G^-(\cdot)$ with attribute strength $+\delta$ and $-\delta$ to an image $x$, and ensure the resulting image $G^-(G^+(x))$ would arrive close to the input $x$. The circle consistency loss is defined as:
    \begin{equation}
       \begin{split}
           \mathcal{L}_{\text{Cycle}}&=||G^-(G^+(x))-x||_1+||G^+(G^-(y))-y||_1,
        \end{split}
    \end{equation}
where $x$ and $y$ are the training image pair with opposite attribute labels, and we impose the circle consistency constraint for both of them. The $L1$ norm is used to measure the image distance.

\begin{figure*}
	\begin{center}
       \includegraphics[width=2.0\columnwidth]{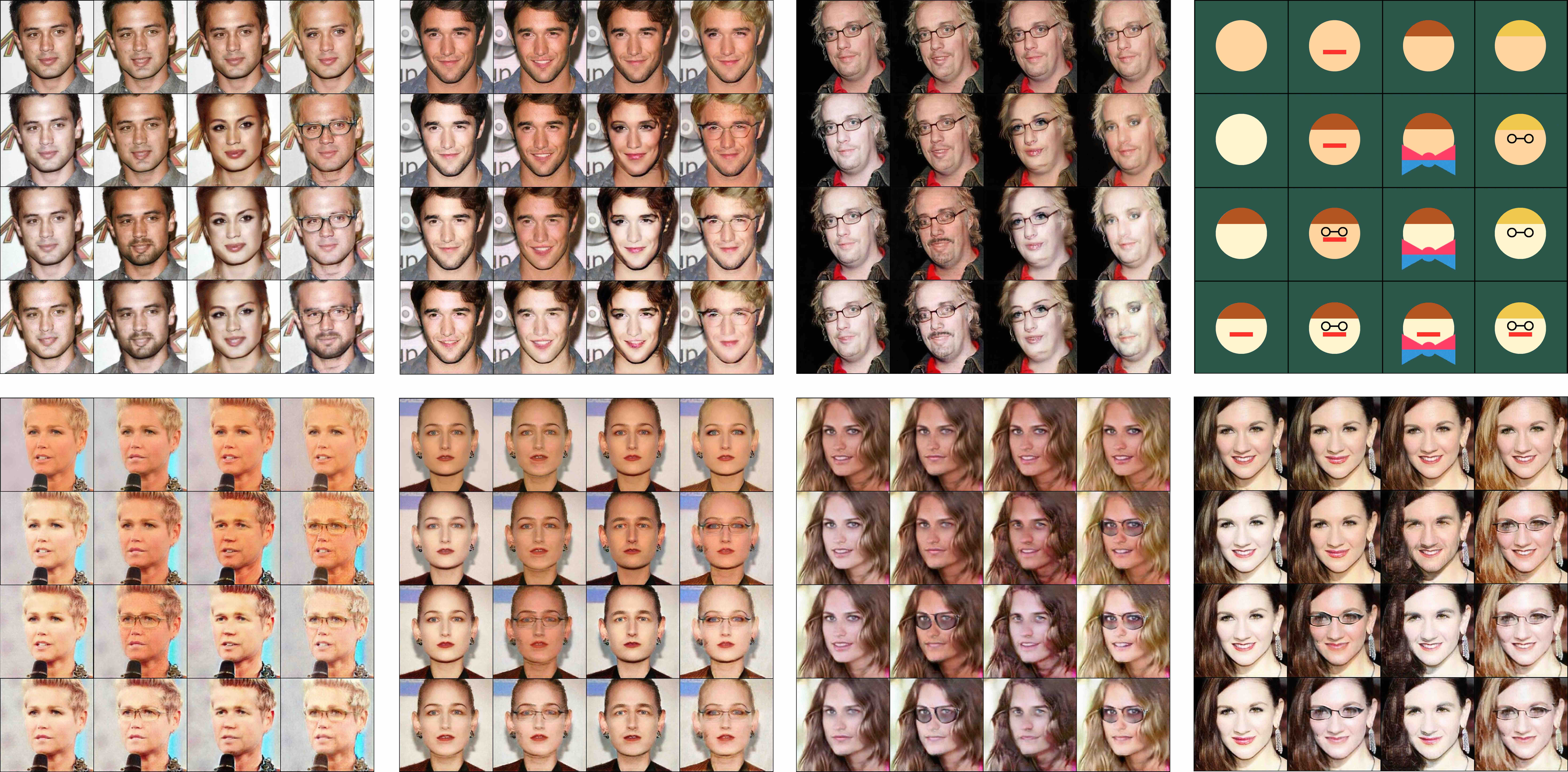}
	\end{center}
	\caption{More results of face attribute manipulation by our M-AAE method. The manipulated attributes for male (first row) are the same as those in Fig. 3, while the manipulated attributes for female (second row) are shown at top-right.}
	\label{Fig.4}
\end{figure*}

\paragraph{Mask loss}

In some cases, we observed the image background would change along with the foreground face by previous attribute manipulation methods. This is not visually pleasing and also suggests some manipulation efforts are wasted in wrong regions. We claim that pasting the original background around the manipulated face is not ideal because the two parts can be incompatible. Here we learn to change the foreground face attribute and keep background the same in a coherent way. We generate a facial mask (thus background mask as well) by using FCN~\cite{Long2015Fully}, and penalize the background difference between input $x$ and generated $G(x)$:
    \begin{equation}
           \mathcal{L}_{\text{Mask}}=||\text{Mask}(G(x)) -\text{Mask}(x)||_1,
    \end{equation}
where $\text{Mask}(\cdot)$ is the mask-out operator using the generated background mask. Note the background mask of input $x$ is shared for both input $x$ and output $G(x)$. We do not generate a separate mask for $G(x)$ which leads to inconsistent penalty.

\subsection{Overall Training Procedure}

Our final training objective is defined as follows:
   \begin{equation}
       \begin{split}
           \min \limits_{G} \max \limits_{D} & \;\;\; \alpha_1 \mathcal{L}_{\text{VAE}} + \alpha_2 \mathcal{L}_{\text{GAN}} \\& + \alpha_3  \mathcal{L}_{\text{ID}} + \alpha_4 \mathcal{L}_{\text{Cycle}} + \alpha_5 \mathcal{L}_{\text{Mask}},
       \end{split}
       \label{eq6}
   \end{equation}
where the weights of $\alpha_1 \sim \alpha_5$ balance the relative importance of our 5 loss terms. The GAN generator,~i.e.,~the encoder-decoder are trained jointly, while the GAN discriminator is trained alternatively. The face recognition network is only used to extract features and its weights are frozen. We choose the first 11 layers of the recognition network~\cite{Parkhi2015Deep} as feature extractor.

\section{Experiments}

In this section, we first introduce our used dataset and implementation details. Our M-AAE is compared against state-of-the-arts both qualitatively and quantitatively to show our advantage. Ablation study is conducted to demonstrate the contribution of each component of our framework.

\subsection{Dataset and Implementation Details}

We evaluated on the CelebA dataset~\cite{liu2015faceattributes}. This dataset contains 202599 face images of 10177 celebrities. Each image is labeled with 40 binary attributes,~e.g.,~"hair color", "age", "gender" and "pale skin". We choose 7 typical attributes (see Fig.~\ref{Fig.3}) for our attribute manipulation experiments. For each attribute, we select 1000 testing images and train with the remaining images in the dataset.

\subsection{Qualitative Evaluation}

Fig.~\ref{Fig.3} compares our M-AAE method qualitatively with the state-of-the-art residual image GAN~\cite{Shen2017Learning}, UNIT~\cite{Liu2017Unsupervised} and StarGAN~\cite{Choi2017StarGAN} in the first row. The recent residual image GAN and StarGAN achieve top performance in image translation and attribute manipulation. The UNIT method is similar to ours in using the VAE-GAN framework and cycle-consistency constraint. We observed that all these methods can produce artifacts or lose personal features to some extent. Their performance is usually good on single attribute manipulation or multi-attribute manipulation when the target attributes are correlated (e.g.,~"pale skin" and "gender"). However, the performance deteriorates in more complex scenarios. For example, residual image GAN totally collapses while generating images with eyeglass. In comparison, our M-AAE method (rightmost, bottom row) consistently produces photorealistic and faithful images with different attributes.

\noindent
{\bf Ablation Study} Fig.~\ref{Fig.3} also compares our various baselines to demonstrate the contribution of our major components. From the comparison of results in (e) and (f), we can find that modifying a meaningful subset of feature map pixels can better preserve global face information (e.g.,~color tone) than modifying the entire feature map. Note the two baselines already use the cycle consistency loss in our VAE-GAN framework, whose efficacy is validated by similar works like UNIT~\cite{Liu2017Unsupervised}. Hence in (g), we further show that adding an ID loss can enhance the identify preservation while editing other attributes. When we use an extra mask loss, the background is made sharper and the foreground facial details also get enhanced with higher fidelity.

\begin{table}[t]
\caption{The perceptual evaluation for ranking different methods on the multi-attribute manipulation task on CelebA. The average rank (between 1 and 7, from best to worst) is shown in each case. The top cell compares state-of-the-art methods, while the bottom cell compares several baselines of ours.}
\centering
\resizebox{1.0\columnwidth}{!}{
  \begin{tabular}{|c|c|c|c|c|}  
    \hline  
    Num of manipulated attributes &  1  &  2 &  3 &  4\\
    \hline  
    \hline 
    Residual image GAN & 6.09 & 6.36 & 5.94 & 6.70  \\   
    \hline  
    UNIT & 3.93 & 5.85 & 5.62 & 5.71 \\   
    \hline 
    StarGAN & 5.16 & 4.40 & 4.24 & 5.07 \\   
    \hline 
    \hline 
    Modify entire feature map & 3.75 & 3.83 & 4.43 & 3.87  \\ 
    \hline 
    Modify feature map sparsely & 3.02 & 2.75 & 2.66 & 2.84  \\  
    \hline 
    + ID loss & 2.72 & 2.57 & 2.61 & 2.49 \\  
    \hline 
    + ID loss + Mask loss (Ours) & 2.39 & 1.79 & 1.70 & 2.10  \\     
    \hline 
  \end{tabular}
  }
  \label{table2}   
\end{table}

\subsection{Quantitative Evaluation}
For quantitative evaluations, we perform a user study inviting volunteers to evaluate the attribute manipulation results. Given a set of generated images from different methods, the volunteers are instructed to rank the methods based on perceptual realism, quality of transfered attribute and preservation of personal features. The generated images from different methods are shuffled before presented. There are 30 validated volunteers to evaluate results with the 7 attributes chosen from CelebA. The average rank (between 1 and 7) of each method is calculated and shown in Table~\ref{table2}. Note that we experiment with different numbers of manipulated attributes from 1 to 4, which have gradually increasing difficulty. From the shown results, again, we find our advantage over prior methods (ranks higher), especially in the multi-attribute manipulation cases. Our ID loss and Mask loss help to improve the results steadily due to their preservation of foreground facial details and background scene.

 \begin{figure}
	\begin{center}
       \includegraphics[width=1.0\columnwidth]{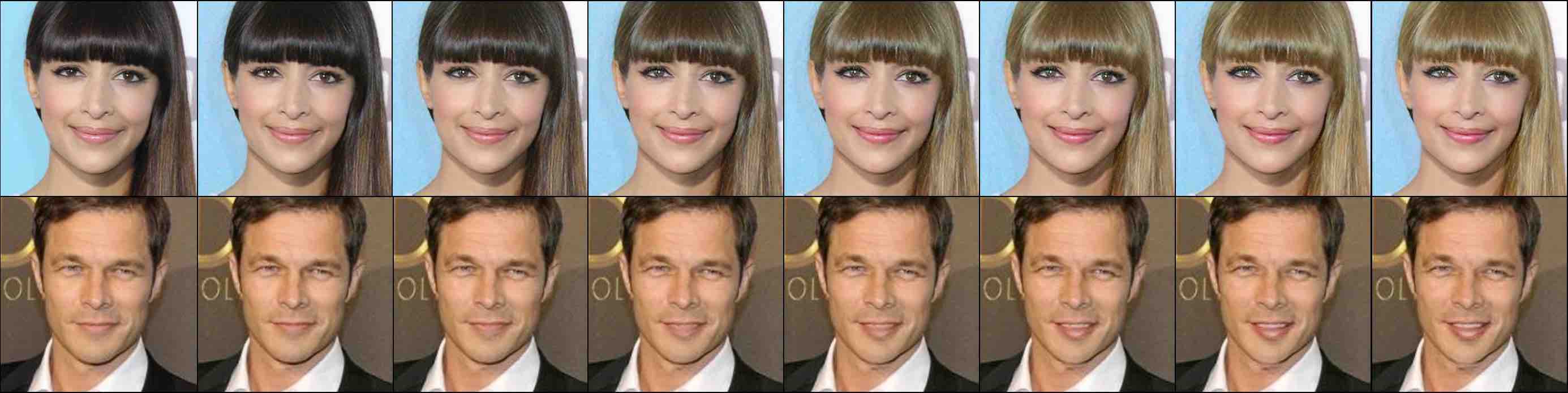}
	\end{center}
	\caption{Continuous manipulation of attributes of blond hair (first row) and mouth open (second row) by our method.}
	\label{Fig.5}
\end{figure}

\subsection{Analysis}
We show more of our results in Fig.~\ref{Fig.4} to empirically prove the generalization ability of our method. Our method handles well with a rich combination of attributes, successfully preserving the unique facial details and background in the generated image with a different attribute. We also show our capability of continuous manipulation of attribute strength in Fig.~\ref{Fig.5}. We achieve this by adjusting the attribute strength in latent features, which is more favorable than prior methods that take a fixed attribute vector as input.

\section{Conclusion and Future Work}

In this paper, we propose a Mask-Adversarial AutoEncoder (M-AAE) method to effectively manipulate human face attributes. Our method is based on the VAE-GAN framework, and we propose an effective method to modify a minimum number of pixels in the feature maps of an encoder, which allows us to change the attribute strength continuously without hindering global information. Our method pays special attention to facial detail preservation and image background consistency. We introduce the face recognition loss and cycle consistency loss for faithful preservation of face details, and also propose a mask loss to ensure background consistency. Experiments show that our method can generate highly photorealistic and faithful images with changing attributes. In principle, our method can be extended to deal with more image translation tasks (e.g., style transformation) which will be included in our future work.

\appendix

\bibliographystyle{named}
\bibliography{ijcai18}

\end{document}